\documentclass{article}
\usepackage{ijcai15}
\usepackage{times}
\usepackage{color}
\usepackage{amsmath}
\usepackage{amsfonts}
\usepackage{url}
\usepackage{ntheorem}
\usepackage{graphicx}
\usepackage{multirow}

\newcommand{\oldsolver}{\texttt{sunny-cp}}
\newcommand{\solver}{\oldsolver\texttt{2}}

\newcommand{\all}{h_{all}}
\newcommand{\sch}{h_{sch}}
\newcommand{\sel}{h_{sel}}

\newcommand{\ps}[2]{\mathbb{P}_{#1, #2}}
\newcommand{\rnk}{\mathit{rank}_\sigma}
\newcommand{\vpar}{\text{VPS}}

\newcommand{\proven}{\texttt{proven}}
\newcommand{\otime}{\texttt{time}}
\newcommand{\score}{\texttt{score}}
\newcommand{\area}{\texttt{area}}
\newcommand{\sunny}[1]{\solver_{[#1]}}

\usepackage{times}
\title{A Multicore Tool for Constraint Solving}
\author{Roberto Amadini, Maurizio Gabbrielli, Jacopo Mauro \\
 Department of Computer Science and Engineering, University of Bologna / Lab. Focus INRIA \\
\{amadini,gabbri,jmauro\}@cs.unibo.it
}

\begin{document}

\maketitle

\begin{abstract}
In Constraint Programming (CP), a portfolio solver uses a 
variety of different solvers for solving a given 
Constraint Satisfaction\,/\,Optimization Problem. 
In this paper we introduce \solver{}: the first parallel CP portfolio solver that
enables a dynamic, cooperative, and 
simultaneous execution of its 
solvers in a multicore setting.
It incorporates state-of-the-art solvers, providing also a 
usable and configurable framework.
Empirical results are very promising. 
\solver{} can even outperform the 
performance of the oracle solver which always selects the best solver of the portfolio for a 
given problem.
\end{abstract}

\section{Introduction}
The \emph{Constraint Programming} (CP) paradigm enables to express complex 
relations in form of constraints to be satisfied. 
CP allows to model and solve \emph{Constraint Satisfaction Problems} 
(CSPs) as well as \emph{Constraint Optimization Problems} (COPs)~\cite{handbook}.
Solving a CSP means finding a solution that satisfies all the constraints 
of the problem, while a COP is a generalized CSP where the goal is 
to find a solution that minimises or maximises an objective function. 

A fairly recent trend in CP is solving a problem by means of portfolio 
approaches~\cite{DBLP:journals/ai/GomesS01}, which can be seen 
as instances of the more general Algorithm Selection problem~\cite{rice_algorithm_selection}. 
A \textit{portfolio solver} is a particular solver that exploits a collection 
of $m > 1$ constituent solvers $s_1, \dots, s_m$ to 
get a globally better solver. When a new, unseen problem $p$ comes, the 
portfolio solver tries to predict the best constituent solver(s) 
$s_{i_1}, \dots, s_{i_k}$ (with $1 \leq k \leq m$) for solving $p$ and then runs them.

Unfortunately ---despite their proven effectiveness--- portfolio solvers are 
scarcely used outside the walls of solvers competitions, and  
often confined to SAT solving. Regarding the CP field, the first 
portfolio solver (for solving CSPs only) was CPHydra~\cite{cphydra} that in 2008 
won the International CSP Solver Competition.
In 2014, the sequential portfolio solver \oldsolver{}~\cite{sunnycp} attended the
\emph{MiniZinc Challenge} (MZC)~\cite{DBLP:journals/constraints/StuckeyBF10}  
with respectable results 
(4\textsuperscript{th} out of 18). 
To the best of our knowledge, no similar tool exist for solving both CSPs and COPs.


In this paper we take a major step forward by introducing \solver{}, a 
significant enhancement of \oldsolver{}. Improvements are manifold. Firstly, 
\solver{} is \emph{parallel}: it exploits 
multicore architectures for (possibly) running more constituent solvers simultaneously.
Moreover, while most of the parallel portfolio solvers are static (i.e., 
they decide off-line the solvers to run, regardless of the problem to be solved),  
\solver{} is instead 
\emph{dynamic:} the solvers scheduling is predicted on-line according to a 
generalization of the SUNNY algorithm~\cite{sunny}.
Furthermore, for COPs, the parallel execution is also \emph{cooperative}: a running 
solver can exploit the current best bound found by another one through a 
configurable \emph{restarting} mechanism.
Finally, \solver{} enriches \oldsolver{} by incorporating new, 
state-of-the-art solvers and by providing a framework which is more 
\emph{usable} and \emph{configurable} for the end user.

We validated \solver{} on two benchmarks of CSPs and COPs (about 5000 
instances each). Empirical evidences show very promising results: its performance 
is very close (and sometimes even better) to that of the \emph{Virtual Best Solver} (VBS), i.e.,
the oracle solver which always selects the best solver of the portfolio for a given problem.
Moreover, for COPs, 
\solver{} using $c = 1,2,4,8$ cores always outperforms the \emph{Virtual $c$-Parallel 
Solver} ($\vpar_c$), i.e., a static 
portfolio 
solver that runs in parallel a fixed selection of the best
$c$ solvers of the portfolio.

\emph{Paper Structure.} In Section \ref{sec:sunny} we generalize 
the SUNNY algorithm for a multicore setting. In Section \ref{sec:solver} we describe 
the architecture and the novelties of \solver{}, while in Section 
\ref{sec:validation} we discuss its empirical evaluation. 
In Section \ref{sec:related} we report the related work before concluding in 
Section \ref{sec:concl}.


\section{Generalizing the SUNNY Algorithm}
\label{sec:sunny}
SUNNY is the algorithm that underpins \oldsolver{}.
Fixed a solving timeout $T$ and a portfolio $\Pi$, SUNNY exploits 
instances similarity to produce a 
sequential schedule $\sigma = [(s_1, t_1), \dots, (s_n, t_n)]$ where solver
$s_i$ has to be run for $t_i$ seconds and $\sum_{i = 1}^{n} t_i = T$. 
For any input problem $p$, SUNNY uses a \emph{$k$-Nearest Neighbours} 
($k$-NN) algorithm to select from a training set of known instances the subset 
$N(p, k)$ of the $k$ instances closer to $p$.
According to the $N(p, k)$ instances, SUNNY relies on three heuristics:
$\sel$, for \emph{selecting} the most promising solvers to run; $\all$, 
for \emph{allocating} to each solver a certain runtime (the more a solver is 
promising, the more time is allocated); and $\sch$, for \emph{scheduling} the 
sequential execution of the solvers according to their presumed speed.
These heuristics depend on the application domain. For example, for 
CSPs $\sel$ selects the smallest sub-portfolio $S \subseteq \Pi$ that solves the 
most instances in $N(p, k)$, by using the solving time for breaking ties.
$\all$ allocates to each $s_i \in S$ a time $t_i$ proportional to 
the instances that $S$ can solve in $N(p, k)$, while $\sch$ sorts 
the solvers by increasing solving time in $N(p, k)$. For COPs the approach is 
analogous, but different performance metrics are used~\cite{paper_lion}.


As an example, let
$p$ be a CSP, $\Pi = \{s_1, s_2, s_3, s_4 \}$ a portfolio of 4 solvers, $T = 1800$ seconds   
the solving timeout, $N(p, k) = \{p_1, ..., p_4\}$ the $k = 4$ neighbours of $p$, 
and the runtimes of solver $s_i$ on problem $p_j$ defined as in Table 
\ref{table:runtimes}.
In this case, the smallest sub-portfolios that solve the most instances are $\{s_1, s_2, s_3\}$, 
$\{s_1, s_2, s_4\}$, and $\{s_2, s_3, s_4\}$. The heuristic $\sel$ selects $S = \{s_1, s_2, s_4\}$ 
because these solvers are faster in solving the instances in the neighbourhood. Since $s_1$ and 
$s_4$ solve 2 instances and 
$s_2$ solves 1 instance, $\all$ splits the solving time window $[0, T]$ into $5$ slots: 
2 allocated to $s_1$ and $s_4$, and 1 to $s_2$. Finally, $\sch$ sorts the solvers 
by increasing solving time. The final schedule produced by SUNNY is therefore 
$\sigma = [(s_4, 720), (s_1, 720), (s_2, 360)]$.
\begin{table}[ht]
\centering
\scalebox{0.9}{
  \begin{tabular}{|c||c|c|c|c|}
    \cline{2-5}
    \multicolumn{1}{c||}{} & $p_1$ & $p_2$ & $p_3$ & $p_4$ \\
    \cline{1-5}
    $s_1$ & \emph{T} & \textbf{3} & \emph{T} & \textbf{278} \\
    \cline{1-5}
    $s_2$ &  \textbf{593} & \emph{T} & \emph{T} & \emph{T} \\
    \cline{1-5}
    $s_3$ & \emph{T} &  \textbf{36} & \textbf{1452} & \emph{T} \\
    \cline{1-5}
    $s_4$ & \emph{T} & \emph{T} & \textbf{122} & \textbf{60} \\
    \cline{1-5}
    \multicolumn{1}{c}{}
    \end{tabular}}
\caption{Runtimes (in seconds). $T$ means the solver timeout.}
\label{table:runtimes}
\end{table}

For a detailed description and evaluation of SUNNY we refer the reader to 
\cite{sunny,paper_lion}.
Here we focus instead on how we generalized SUNNY for 
a multicore setting where $c \geq 1$ cores are available and, typically, 
we have more solvers than cores.
\solver{} uses a 
\emph{dynamic} approach: the schedule is decided on-line according 
to the instance to be solved.
The approach used is rather simple: starting from the sequential 
schedule $\sigma$ produced by SUNNY on a given 
problem, 
we first detect the $c-1$ solvers of $\sigma$ having the largest allocated 
time, and we assign a different core to each of them. The other solvers of 
$\sigma$ (if any) are scheduled on the last core, by preserving their 
original order in $\sigma$ and by widening their allocated times to cover the 
entire time window $[0, T]$.

Formally, given $c$ cores and the sequential schedule $\sigma = [(s_1, t_1), \dots, (s_n, t_n)]$ 
computed by SUNNY, 
we define a total order $\prec_\sigma$ such that 
$s_i \prec_\sigma s_j \Leftrightarrow t_i > t_j \vee (t_i = t_j \wedge i < j)$, 
and a ranking function $\rnk$ such that $\rnk(s) = i$ if and only if $s$ is the $i$-th solver of 
$\sigma$ according to $\prec_\sigma$.
We define the 
\emph{parallelisation} of $\sigma$ on $c$ cores as a function $\ps{\sigma}{c}$ 
which associates to each core $i \in \{1, \dots, c\}$ a sequential schedule 
$\ps{\sigma}{c}(i)$ such that:
\[
\scalebox{0.96}{$
\ps{\sigma}{c}(i) =
  \begin{cases}
  [~] \hspace*{153pt} \text{if }  i > n \\
  [ (s, T) ] \hspace*{47pt} \text{if } i = \rnk(s) \wedge 1 \leq i < c \\
  [ (s, \frac{T}{T_c} t) ~|~ (s, t) \in \sigma \wedge \rnk(s) \geq c] \hspace*{11pt} \text{if } i = c \\
\end{cases}$}
\]
where $T_c = \sum\{t  ~|~ (s,t) \in \sigma \wedge \rnk(s) \geq c\}$.

Let us consider the schedule of the previous example:
$\sigma = [(s_4, 720), (s_1, 720), (s_2, 360)]$. With 
$c = 2$ cores, the  
parallelisation of $\sigma$ is defined as $\ps{\sigma}{2}(1) = [(s_4, 1800)]$ and 
$\ps{\sigma}{2}(2) = [(s_1, 1800/1080 \times 720), (s_2, 1800/1080 \times 360)]$.

The rationale behind the schedule parallelisation $\ps{\sigma}{c}$ is that SUNNY aims to 
allocate more time to the most promising solvers, scheduling them according to 
their speed. 
Assuming a good starting schedule $\sigma$, its parallelisation 
$\ps{\sigma}{c}$ is rather straightforward and
it is easy to prove that, assuming an independent execution of 
the solvers without synchronization and memory contention issues, 
$\ps{\sigma}{c}$ can never solve less problems than $\sigma$.

Obviously, different ways to parallelise the sequential schedule are possible. 
Here we focus on what we think to be one of the most simple yet promising ones. 
An empirical 
comparison 
of other parallelisation methods is outside the scope of this paper.

%



\section{SUNNY-CP 2}
\label{sec:solver}
\solver{} solves both CSPs and COPs encoded in the \emph{MiniZinc} 
language~\cite{Nethercote07minizinc:towards}, 
nowadays the de-facto standard for modelling CP problems. By default, \solver{} uses 
a portfolio $\Pi$ of 12 solvers disparate in their nature. In 
addition to the 8 solvers of \oldsolver{} (viz., Chuffed, CPX, G12/CBC, G12/FD, 
G12/LazyFD, G12/Gurobi, Gecode, and MinisatID) it includes new, 
state-of-the-art solvers able to win four gold medals in the 
MZC 2014, namely: Choco, HaifaCSP, iZplus, and OR-Tools.

\begin{figure}[ht]
\centering
\includegraphics[width=0.34\textwidth]{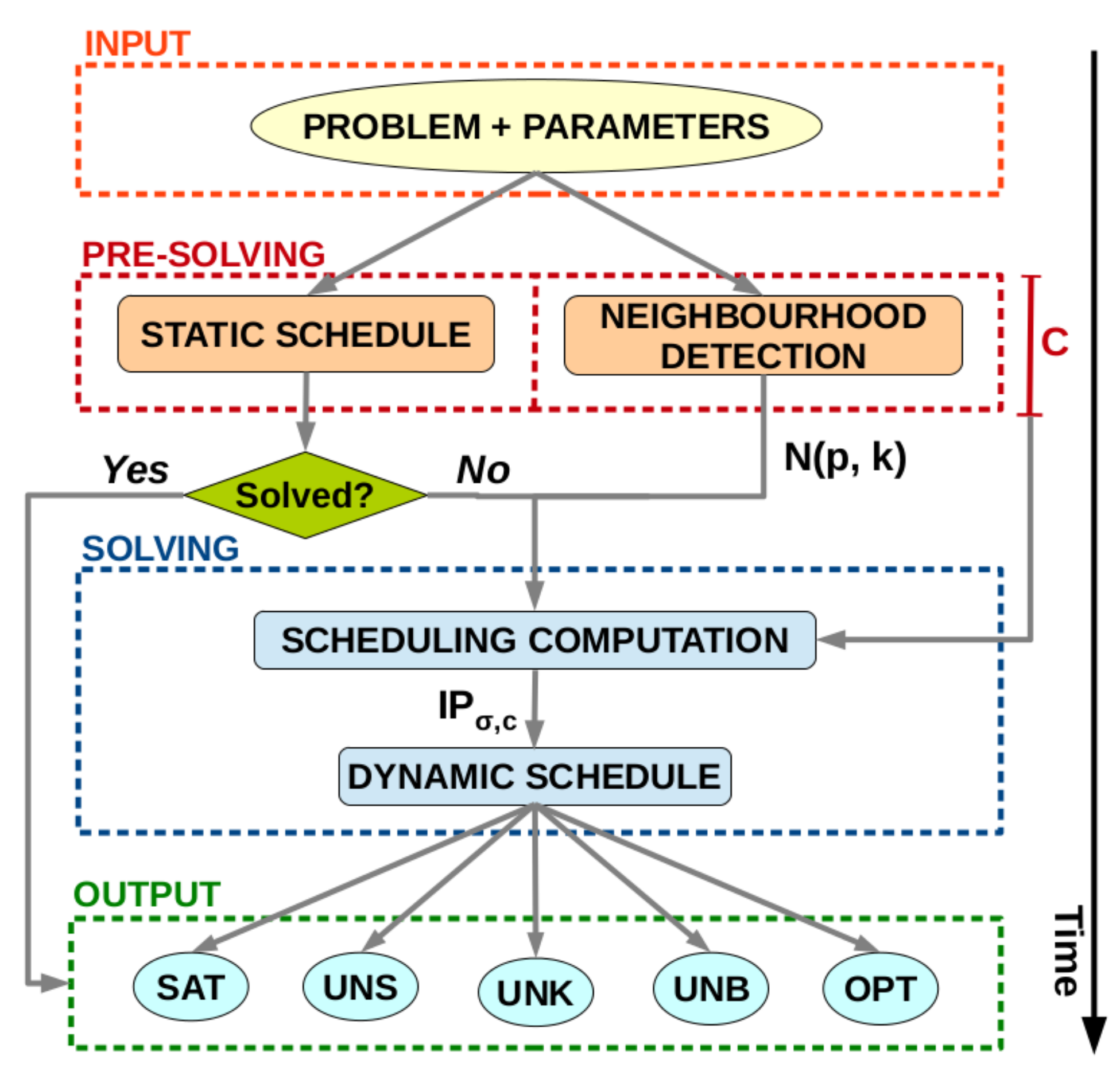}
\caption{$\solver$ architecture.}
\label{fig:solver}
\end{figure}
Figure \ref{fig:solver} summarizes the execution flow of \solver{}. 
It takes as input a problem $p$ to be solved and a list of input parameters 
specified by the user (e.g., the timeout $T$ used by SUNNY, the 
number of cores $c$ to use, etc.). If a parameter is not 
specified, a corresponding default value is used.

The solving process then relies on two sequential steps, later detailed: 
\emph{(i)} the \emph{pre-solving} phase, where a static schedule of solvers 
may be run and the instance neighbourhood is computed; \emph{(ii)} the 
\emph{solving} phase, which runs the dynamic schedule on different cores. 

If $p$ is a CSP, the output of \solver{} can be either 
\emph{SAT} (a solution exists for $p$), 
\emph{UNS} ($p$ has no solutions), or \emph{UNK} (\solver{} can not say 
anything about $p$). In addition, if $p$ is a COP, two more answers are possible: 
\emph{OPT} (\solver{} proves the optimality of a solution), or 
\emph{UNB} ($p$ is unbounded).

\subsection{Pre-solving}
\label{sec:presolve}
The pre-solving step consists in the 
simultaneous execution on $c$ cores of two distinct tasks: the execution of a 
(possibly empty) static schedule, and the neighbourhood detection.

Running for a short time a static schedule has proven to be effective. It 
enables to solve 
those instances that some solvers can solve very quickly (e.g., see \cite{3s}) and, 
for COPs, to quickly find good sub-optimal solutions~\cite{paper_cp}.

The \textit{static schedule} $\overline{\sigma} = [(s_1, t_1), \dots, (s_n, t_n)]$ 
is a prefixed schedule of $n \geq 0$ solvers decided off-line, 
regardless of the input problem $p$. To execute it, \solver{}
applies a ``First-Come, First-Served'' policy by following the schedule order.
Formally, the first $m$ solvers $s_1, \dots, s_m$ with  $m = \min(c, n)$ are launched on 
different cores. Then, for $i = m+1, ..., n$, the solver $s_{i}$ is started as soon 
as a solver $s_j$ (with $1 \leq j \leq i$) terminates its execution without 
solving $p$.
If a solver $s_i$ fails prematurely before 
$t_i$ (e.g., due to memory overflows or unsupported constraints) then $s_i$ is 
discarded from the portfolio for avoiding to run it again later. If instead $s_i$ reached 
its timeout, then it is just suspended: if it has to run 
again later, it will be resumed instead of restarted from scratch.
The user has the possibility of setting the static schedule as an input 
parameter. For simplicity, the static 
schedule of \solver{} is empty by default.

When solving COPs, the timeout defined by the user may be overruled by \solver{} 
that examines the solvers behaviour and may decide to delay the interruption of 
a solver. Indeed,
a significant novelty of \solver{} is the 
use of a \emph{waiting threshold} $T_w$. A solver $s$ scheduled for $t$ 
seconds is never interrupted if it has found a new solution in the last 
$T_w$ seconds, regardless of whether the timeout $t$ is expired or not. 
The underlying logic of $T_w$ is to not stop a solver which is actively 
producing solutions. Thus, if $T_w > 0$ a solver may be executed for more than 
its allocated time. Setting $T_w$ to a high value might therefore delay, and 
thus hinder, the execution of the other solvers.
%

For COPs, \solver{} also enables the \emph{bounds communication} between solvers:
the sub-optimal solutions found by a solver are used to narrow the search space of 
the other scheduled solvers. In \cite{paper_cp} this technique is successfully 
applied in a sequential setting, where the best objective bound found by a solver 
is exploited by the next scheduled ones. Things get more complicated in a 
multicore setting, where solvers are run simultaneously as black boxes and 
there is no support for 
communicating bounds to another running solver without interrupting its solving 
process.
We decided to overcome this problem by using a 
\emph{restarting threshold} $T_r$. 
If a running solver 
$s$ finds no solution in the last $T_r$ seconds, and its current best 
bound $v$ is obsolete w.r.t.~the overall best bound $v'$ found by another solver, 
then $s$ is restarted with the new bound $v'$.
The reason behind $T_r$ is that restarting an active solver is 
risky and potentially harmful even when its current best bound is actually 
obsolete. However, 
also ignoring the solutions found by other solvers could mean to neglect valuable 
information. The choice of $T_r$ is hence critical: a too small value might 
cause too many restarts, while a big threshold inhibits the bound communication.
Care should be taken also because restarting a solver means to lose all the 
knowledge it gained during the search.

%

After performing some empirical investigations, we found reasonable to set 
the default values of $T_w$ and $T_r$ to 2 and 5 seconds respectively. The 
user can however configure such 
parameters as she wishes, even by defining a customized setting for each different 
solver of the portfolio.

The \textit{neighbourhood detection} phase begins in parallel with the static schedule 
$\overline{\sigma}$ execution 
only after all the solvers of $\overline{\sigma}$ has been started.
This phase has lower priority than $\overline{\sigma}$ since its purpose is not quickly 
finding solutions, but detecting the neighbourhood 
$N(p, k)$ later on used to compute the dynamic schedule. For this reason, the computation of 
the neighbourhood starts as soon as one core is free (i.e., the number of running solvers is 
smaller than $c$). 
The first step of pre-scheduling is the extraction of the \emph{feature 
vector} of $p$, i.e., a collection of numerical attributes 
(e.g., number of variables, of constraints, etc.) that characterizes the 
instance $p$.
The feature vector is then used for detecting the set $N(p, k)$ of the 
$k$ nearest neighbours of $p$ within a dataset of known instances. 

The default dataset $\Delta$ of \solver{} is the union of a set $\Delta_\text{CSP}$ of 
5527 CSPs and a set $\Delta_\text{COP}$ of 4988 COPs, retrieved from the instances of the 
MiniZinc 1.6 benchmarks, the MZCs 2012--2014, and the International CSP Solver Competitions 
2008/09.
\solver{} provides a default knowledge base that associates to every 
instance $p_i \in \Delta$ its feature vector and the runtime information 
of each solver of the portfolio on $p_i$ (e.g., solving time, 
anytime performance, etc.). In 
particular, the feature vectors are used for computing 
$N(p, k)$ while the runtime information are later used for computing the dynamic 
schedule $\sigma$ by means of SUNNY algorithm.

The feature extractor used by \solver{} is the same as \oldsolver{}.\footnote{
The feature extractor \texttt{mzn2feat} is available 
at \url{https://github.com/CP-Unibo/mzn2feat}.
For further information please see  
\cite{mzn2feat} and \cite{sunnycp}.}
The neighbourhood size is set by default to $k = 70$: following \cite{knn}
we chose a value of $k$ close to $\sqrt{n}$, where $n$ is the number of 
training samples.
Note that during the pre-solving phase only $N(p, k)$ is computed.
This is because SUNNY requires a timeout $T$ for computing the schedule $\sigma$. 
The total time taken by the pre-solving phase ($C$ in Figure \ref{fig:solver}) 
must therefore be subtracted from the initial timeout $T$ (by default,  
$T = 1800$ seconds as in \oldsolver). This can be done only when the pre-solving 
ends, since $C$ is not predictable in advance.


The pre-solving phase terminates when either: \emph{(i)} a solver of the 
static schedule $\overline{\sigma}$ solves $p$, or \emph{(ii)} the 
neighbourhood computation and all the solvers of $\overline{\sigma}$ have finished their 
execution. In the first case \solver{} outputs the solving outcome and stops its execution, 
skipping the solving phase since the instance is already solved.

\subsection{Solving}
The solving phase receives in input the time $C$ taken by pre-solving and the 
neighbourhood $N(p, k)$. These parameters are used by SUNNY to dynamically 
compute the sequential schedule $\sigma$ with a reduced timeout $T - C$. 
Then, $\sigma$ is parallelised according to the $\ps{\sigma}{c}$ operator 
defined in Section \ref{sec:sunny}.
Once computed $\ps{\sigma}{c}$, each scheduled solver is run on $p$ 
according to the input parameters. 
If a solver of $\ps{\sigma}{c}$ 
had already been run in the static schedule $\overline{\sigma}$, then its 
execution is resumed instead of restarted. If $p$ is a COP, the solvers 
execution follows the approach explained in Section \ref{sec:presolve} according 
to $T_w$ and $T_r$ thresholds. As soon as a solver of 
$\sigma$ solves $p$, the execution of \solver{} is interrupted and the solving 
outcome is outputted.

Note that we decided to make \solver{} an \emph{anytime solver}, i.e., a solver that can run 
indefinitely until a solution is 
found. For this reason we let the solvers run 
indefinitely even after the timeout $T$  until a solution is found.
Moreover, at any time we try to have exactly $c$ running solvers. 
In particular, if $c$ is greater or equal than the portfolio size no prediction is 
needed and we simply run a solver per core for indefinite time. 
When a scheduled solver fails during its execution, the overall solving 
process is not affected since \solver{} will simply run the next scheduled solver, 
if any, or another one in its portfolio otherwise.

Apart from the scheduling parallelisation, \solver{} definitely improves 
\oldsolver{} also from the engineering point of view. 
One of the main purposes of 
\solver{} is indeed to provide a framework which is easy to use and configure by the end 
user. First of all, \solver{} provides utilities and procedures for adding new 
constituent solvers to $\Pi$ and for customizing their settings. 
Even though the feature extractor of 
\solver{} is the same as \oldsolver, the user can now
define and use her own tool for feature processing.
The user can also define her own knowledge base, starting from raw 
comma-separated value files, 
by means of suitable utility scripts. Furthermore, \solver{} is much more 
parametrisable than \oldsolver. Apart from the aforementioned $T_w$ and $T_r$ 
parameters, 
\solver{} provides many more options allowing, e.g., to set the 
number $c$ of cores to use, limit the memory usage, ignore the search annotations, and specify 
customized options for each constituent solver.

The source code of \solver{} is entirely written in Python and publicly 
available at \url{https://github.com/CP-Unibo/sunny-cp}.

\section{Validation}
\label{sec:validation}

\begin{table*}
\centering
\begin{tabular}{|c|c|c|c|c|c|c|c|c|c|c|}
\hline
\multirow{2}{*}{Metric} & \multirow{2}{*}{\oldsolver{}} & \multicolumn{4}{|c|}{\solver{}} & 
\multicolumn{4}{|c|}{\vpar{}} & \multirow{2}{*}{VBS} \\
\cline{3-10}
 & & 1 core & 2 cores & 4 cores & 8 cores & 1 core & 2 cores & 4 cores & 8 cores & \\
\hline
\proven{} (\%) & 83.26 & 95.26 & 99.11 & \textbf{99.38} & 99.35 & 89.04 & 93.51 & 95.19 & 
99.24 & \textbf{100} \\
\otime{} (s)& 504.10 & 223.21 & 136.04 & 112.60 & \textbf{112.32} & 297.30 & 211.13 & 176.81 & 
\textbf{68.52} & \textbf{54.30}\\
\hline
\end{tabular}
\caption{Experimental results over CSPs.}
\label{table:results_csp}
\end{table*}
\begin{table*}
\centering
\begin{tabular}{|c|c|c|c|c|c|c|c|c|c|c|}
\hline
\multirow{2}{*}{Metric} & \multirow{2}{*}{\oldsolver{}} & \multicolumn{4}{|c|}{\solver{}} & 
\multicolumn{4}{|c|}{\vpar{}} & \multirow{2}{*}{VBS} \\
\cline{3-10}
 & & 1 core & 2 cores & 4 cores & 8 cores & 1 core & 2 cores & 4 cores & 8 cores & \\
\hline
\proven{} (\%) & 71.55 & 74.40 & 74.94 & 75.68 & \textbf{76.34} & 61.33 & 63.45 & 65.60 & 
75.86 & \textbf{76.30}\\
\otime{} (s)& 594.79 & 501.35 & 482.95 & 469.74 & \textbf{454.54} & 718.86 & 682.35 & 645.68 & 
463.63 & 
\textbf{457.00}\\
$\score{} \times 100$ & 90.50 & 92.26 & 93.00 & 93.45 & \textbf{93.62} & 82.80 & 85.99 & 89.53 & 90.79 & \textbf{93.63} \\
\area{} (s) & 257.86 & 197.44 & 149.33 & 138.94 & \textbf{130.53} & 314.07 & 266.11 & 188.20 & 178.81 & \textbf{132.28} \\
\hline 
\end{tabular}
\caption{Experimental results over COPs.}
\label{table:results_cop}
\end{table*}

We validated the performance of \solver{} by running it with its default settings 
on every instance of $\Delta_{\text{CSP}}$ (5527 CSPs) 
and $\Delta_{\text{COP}}$ 
(4988 COPs). Following the standard practices, 
we used a 10-fold cross-validation: we partitioned each dataset in 10 
disjoint folds, treating in turn 
one fold as test set and the union of the 
remaining folds as the training set.


Fixed a solving timeout $T$, different metrics can be adopted to 
measure the effectiveness of a solver $s$ on a given problem $p$. 
If $p$ is a CSP
we typically evaluate whether, and how quickly, $s$ can solve $p$.
For this reason we use two metrics, namely \proven{} and \otime{}, to measure  
the solved instances and the solving time. 
Formally, if $s$ solves 
$p$ 
in $t < T$ seconds, then $\proven(s, p) = 1$ and $\otime(s, p) = t$; otherwise,
$\proven(s, p) = 0$ and $\otime(s, p) = T$.
A straightforward generalization of these two metrics for COPs can be obtained by setting  
$\proven(s, p) = 1$ and $\otime(s, p) = t$ if $s$ proves in $t < T$ seconds the 
optimality of a solution for $p$, the unsatisfiability of $p$ or its unboundedness.
Otherwise, $\proven(s, p) = 0$ and $\otime(s, p) = T$.
However, a major drawback of such a generalization is that the solution 
quality is not addressed.
%
To overcome this limitation, we use in addition the 
\score{} and \area{} metrics introduced in \cite{paper_cp}.

The \score{} metric measures the quality of the best solution found by a solver 
$s$ at 
the stroke of the timeout $T$. More specifically, $\score(s, p)$ is a value in 
$[0.25, 0.75]$ linearly proportional to the distance between the best solution 
that $s$ finds and the best known solution for $p$. An additional reward 
($\score = 1$) 
is given if $s$ is able to prove optimality, while a punishment ($\score = 0$) 
is given if $s$ does not provide any answer.
%
%
Differently from \score{}, the \area{} metric evaluates 
the \emph{behaviour} of a solver in the whole solving time window $[0, T]$ and 
not only at the time edge $T$. It considers the area under the curve that 
associates to each time instant $t_i \in [0, T]$ the best objective 
value $v_i$ found by $s$ in $t_i$ seconds. The 
\area{} value is properly scaled in the range $[0, T]$ so that the more a solver 
is slow in finding good solutions, the more its \area{} is high 
(in particular, $\area = T$ if and only if $\score = 0$).
The \area{} metric folds in a 
number of measures the quality of the best solution found, 
how quickly any solution is found, whether optimality is proven, and how 
quickly good solutions are found.
For a formal definition of \area{} and \score{} we refer the reader 
to \cite{paper_cp}.

We ran \solver{} on all the instances of $\Delta_{\text{CSP}}$ 
and $\Delta_{\text{COP}}$ within a timeout of $T = 1800$ seconds by varying the 
number $c$ of cores in $\{1,2,4,8\}$. We compared \solver{} against 
the very same version of \oldsolver{} submitted to MZC 2014\footnote{The source 
code of \oldsolver{} is publicly available 
at \url{https://github.com/CP-Unibo/sunny-cp/tree/mznc14}
} and the Virtual Best Solver (VBS), i.e., the oracle portfolio solver 
that ---for a given problem and performance metric--- always selects the 
best solver to run. 
Moreover, for each $c \in \{1, 2, 4, 8\}$ we consider as additional baseline 
the \emph{Virtual $c$-Parallel Solver} ($\vpar_c$), i.e., a static portfolio 
solver that always runs in parallel a fixed selection of the best
$c$ solvers of the portfolio $\Pi$.
The $\vpar_c$ is a static portfolio since its solvers are selected in advance, regardless of 
the problem $p$ to be solved. Obviously, the portfolio composition of $\vpar_c$ 
depends on a given evaluation metric. For each $\mu \in \{\proven{}, \otime{}, 
\score{}, \area{}\}$ we therefore defined a corresponding $\vpar_c$ by selecting the 
best $c$ solvers of $\Pi$ according to the average value of $\mu$ over the
$\Delta_{\text{CSP}}$ or $\Delta_{\text{COP}}$ instances.
Note that $\vpar_c$ is an ``ideal'' solver, since it is an \emph{upper bound} of the 
real performance achievable by actually running in parallel all its solvers.
An approach similar to $\vpar_c$ has been successfully used by the SAT 
portfolio solver \texttt{ppfolio}~\cite{ppfolio}.
The $\vpar_1$ is sometimes also called \emph{Single Best 
Solver} (SBS) while $\vpar_{|\Pi|}$ is exactly equivalent to the 
\emph{Virtual Best Solver} (VBS).

In the following, we will 
indicate with $\sunny{c}$ the version of \solver{} exploiting $c$ cores.
Empirical results on CSPs and COPs are summarized in Table 
\ref{table:results_csp} and \ref{table:results_cop} reporting the average values of all the  
aforementioned evaluation metrics.
On average, \solver{} remarkably outperforms all its constituent 
solvers as well as \oldsolver{} for both CSPs and COPs. 
Concerning CSPs, a major boost is given by the introduction of 
HaifaCSP solver in the portfolio.
HaifaCSP solves almost 90\% of $\Delta_{\text{CSP}}$ problems. 
However, as can be seen from the Tables \ref{table:results_csp} and \ref{table:results_cop},  
$\sunny{1}$ solves more than 95\% 
of the instances and, for $c > 1$, $\sunny{c}$ solves nearly all the 
problems. This means that just few cores are enough to almost reach the VBS 
performance. The best $\proven$ performance is achieved by $\sunny{4}$.\footnote{
The practically negligible difference with $\sunny{8}$ is probably due to 
synchronization and memory contention issues (e.g., cache misses) that slow 
down the system when 8 cores are exploited.} Nevertheless, the average 
$\otime$ of $\sunny{8}$ is slightly slower. However, difference are minimal: 
$\sunny{4}$ and $\sunny{8}$ are virtually equivalent.
Note that $\sunny{c}$ is always better than $\vpar_c$ in terms of $\proven$ and,
for $c < 8$, also in terms of $\otime$. Furthermore, $\sunny{1}$ solves more 
instances than $\vpar_4$. In other words, running sequentially the solvers 
of $\solver$ on a 
single core allows to solve more instances than running independently the four 
solvers of the portfolio that solve the most number of problems of $\Delta_{\text{CSP}}$.
Analogously, $\sunny{2}$ solves more instances than $\vpar_4$.
This witnesses that in a multicore setting ---where there are typically more 
available solvers than cores--- 
\solver{} can be very effective.


Even better results are reached by \solver{} on the instances of 
$\Delta_{\text{COP}}$, where there is not a clear dominant solver like 
HaifaCSP for $\Delta_{\text{CSP}}$.
\solver{} outperforms \oldsolver{} in all the considered metrics, and the gain 
of performance in this case is not only due to the introduction of new solvers. 
Indeed, the best solver according to $\otime$ and $\proven$ is Chuffed, which 
was already present in \oldsolver{}. For each $c \in \{1, 2, 4, 8\}$ 
and for each performance metric, $\sunny{c}$ is always better than the 
corresponding $\vpar_c$. Moreover, as for CSPs, \solver{} has very good performances 
even with few cores. For example, by considering the \proven{}, \otime{}, and 
\score{} results, $\sunny{1}$ is always better than $\vpar_4$ and 
its \score{} even outperforms that of $\vpar_8$.
This clearly indicates that the dynamic selection of solvers, together with the 
bounds communication and the waiting/restarting policies implemented by 
\solver{}, makes a remarkable performance difference.

Results also show a nice monotonicity: for all the considered metrics, 
if $c > c'$ then $\sunny{c}$ is better than $\sunny{c'}$ . In particular, 
it is somehow impressive to see that on average \solver{} is able to 
\emph{outperform} the VBS in terms of $\otime$, $\proven$, and $\area$, 
and that for $\score$ the performance difference is negligible (0.01\%).
%
\begin{figure}[t]
\centering
\includegraphics[width=0.48\textwidth]{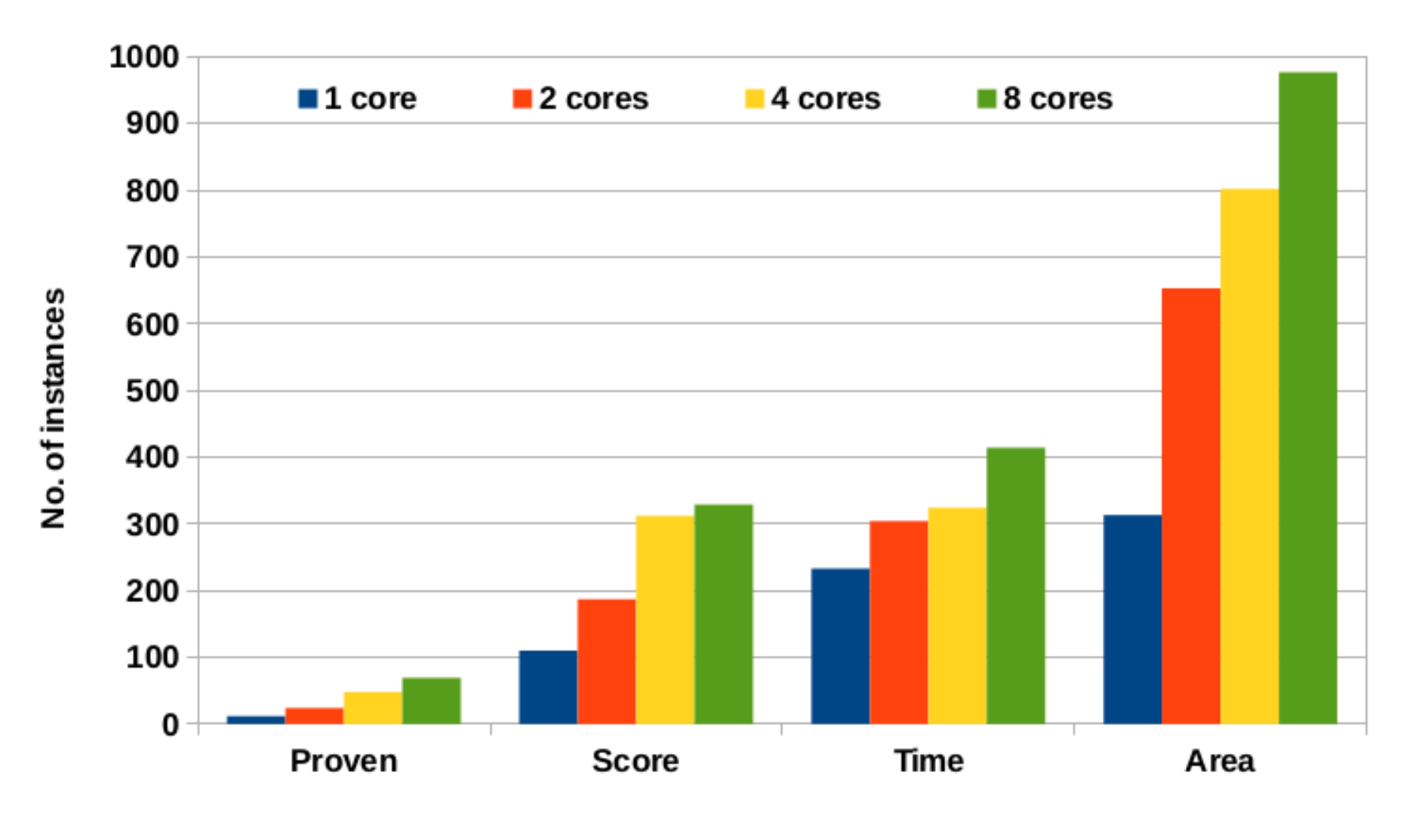}
\caption{Number of COPs for which $\solver{}$ outperforms the VBS.}
\label{fig:betterThanVBS}
\end{figure}
This is better shown in Figure \ref{fig:betterThanVBS} depicting for each 
performance metric the 
number of times that \solver{} is able to overcome the performance of the VBS 
with $c = 1,2,4,8$ cores.
As $c$ increases, the number of times the VBS is beaten gets bigger. In 
particular, the \otime{} and \area{} results show that \solver{} can be very 
effective to reduce the solving time both for completing the search and, 
especially, for quickly finding sub-optimal solutions. For example, for 
414 problems $\sunny{8}$ has a lower \otime{} than VBS, while for 977 instances 
it has a lower \area{}.
Table \ref{table:rcpsp} reports a clear example of the \solver{} potential 
on an instance 
of the Resource-Constrained Project Scheduling Problem (RCPSP)~\cite{brucker1999resource} 
taken from
the MiniZinc-1.6 suite.\footnote{The model is \texttt{rcpsp.mzn} while the 
data is in \texttt{la10\_x2.dzn}}
Firstly, note that just half of the solvers of $\Pi$ can find at least a solution.
Second, these solvers find their best bound $v_{\text{best}}$ very quickly (i.e, 
in a time $t_{\text{best}}$ of between 1.46 and 4.05 seconds) but none of them 
is able to prove the optimality of the best bound $v^* = 958$ within a timeout of 
$T = 1800$ seconds. 
Conversely, \solver{} finds $v^*$ and proves its optimality in less than 11 seconds.
Exploiting the fact that CPX finds $v^*$ in a short time (for CPX 
$t_{\text{best}} = 2.42$ seconds, while \solver{} takes 5 seconds due 
to the neighbourhood detection and scheduling computation), after $T_r = 5$ seconds any other 
scheduled solver is restarted with 
the new bound $v^*$. Now, Gecode can prove almost instantaneously that $v^*$ is optimal, 
while without this help it can not even find it in half 
an hour of computation. 
\begin{table*}
\centering
\begin{tabular}{|c||c|c|c|c|c|c|c||c|}
\cline{2-9}
\multicolumn{1}{c||}{} & Choco & Chuffed & CPX &  G12/FD & Gecode & OR-Tools & Other solvers & \solver{} \\
\hline
$t_{\text{best}}$ (s) & 4.05 & 52.49 & \textbf{2.42} & 2.87 & 3.31 & 1.46 & N/A & 5.00 \\
$v_{\text{best}}$ & 959 & \textbf{958} & \textbf{958} & 959 & 959 & 959 & N/A & \textbf{958} \\
$\otime{}$ & $T$ & $T$ & $T$ & $T$ & $T$ & $T$ & $T$ & \textbf{10.56} \\
\hline
\end{tabular}
\caption{Benefits of \solver{} on a RCPSP instance. $T = 1800$ indicates the solvers timeout.}
\label{table:rcpsp}
\end{table*}

\section{Related Work}
\label{sec:related}

The parallelisation of CP solvers does not appear as 
fruitful as for SAT solvers where techniques like clause sharing are used. As 
an example, in the MZC 2014 the possibility of 
multiprocessing did not lead to remarkable performance gains, despite 
the availability of eight logical cores (in a case, a parallel solver was even 
significantly worse than its sequential version).

Portfolio solvers have proven their effectiveness in many international 
solving competitions. 
For instance, the SAT portfolio solvers 3S~\cite{3s} and CSHC~\cite{cshc} won 
gold medals in SAT Competition 2011 and 2013 respectively. 
SATZilla~\cite{satzilla} won the SAT Challenge 2012, CPHydra~\cite{cphydra} 
the Constraint Solver Competition 2008, the ASP portfolio 
solver claspfolio~\cite{claspfolio} was gold medallist in different tracks of 
the ASP Competition 2009 and 2011, ArvandHerd~\cite{arvandherd} and 
IBaCoP~\cite{ibacop} won some tracks in the 
Planning Competition 2014. 

Surprisingly enough, only a few portfolio solvers are parallel and even 
fewer are the dynamic ones selecting on-line the solvers to run. We are aware 
of only two dynamic and parallel portfolio solvers that attended a solving 
competition, namely p3S~\cite{3s-par} (in the SAT Challenge 2012) and 
IBaCoP2~\cite{ibacop} (in the Planning Competition 2014).
Unfortunately, a comparison of \solver{} with these tools is not possible because  
their source code is not publicly available and they do not deal with COPs.

Apart from the aforementioned CPHydra and \oldsolver{} solvers, 
another portfolio approach for CSPs is Proteus~\cite{proteus}. 
However, with the exception of a preliminary 
investigation about a CPHydra parallelisation \cite{parrallelCSPPortfolio}, 
all these solvers are 
sequential and except for \oldsolver{} they solve just CSPs. Hence, to the best of our knowledge, 
\solver{} is today the only parallel and dynamic CP portfolio solver 
able to deal with also optimization problems.

The parallelisation of portfolio solvers is a hot topic which is drawing some 
attention in the community. For instance, parallel extensions of well-known sequential 
portfolio approaches are studied in \cite{lindauer-lion15a}. In \cite{aspeed} 
ASP techniques are used for computing a static schedule of solvers which can 
even be executed in parallel, while \cite{cire2014parallel} considers the 
problem of parallelising restarted backtrack search for CSPs.
%
%
%

\section{Conclusions}
\label{sec:concl}
In this paper we introduced \solver{}, the first parallel CP portfolio solver 
able to dynamically schedule the solvers to run and to solve both 
CSPs and COPs encoded in the MiniZinc language.
It incorporates state-of-the-art solvers, providing also a 
usable and configurable framework.

The performance of \solver{}, validated on heterogeneous 
and large benchmarks, is promising. 
Indeed, \solver{} greatly outperforms all its constituent solvers and its earlier 
version \oldsolver{}. It can be far better than a ppfolio-like 
approach ~\cite{ppfolio} which statically determine a fixed selection of the best solvers to 
run. For CSPs, \solver{} almost 
reaches the performance of the Virtual Best Solver, while for COPs \solver{} is 
even able to outperform it. 

We hope that \solver{} can stimulate the adoption and the 
dissemination of CP portfolio solvers. Indeed, \oldsolver{} was the only 
portfolio entrant of MiniZinc Challenge 2014. We are interested in submitting \solver{} to the 
2015 edition in order to compare it with 
other possibly parallel portfolio solvers.


There are many lines of research that can be 
explored, both from the scientific and engineering perspective.
As a future work we would like to extend the \solver{} by adding new, possibly parallel 
solvers. Moreover, different parallelisations, 
distance metrics, and 
neighbourhood sizes can be evaluated.

Given the variety of parameters provided by \solver{},
it could be also interesting to exploit Algorithm Configuration techniques~\cite{smac,isac} for 
the automatic tuning of the \solver{} parameters, as well as the parameters of its 
constituent 
solvers.
%

Finally, we are also interested in making \solver{} more usable and portable, e.g., by 
pre-installing it on virtual 
machines or multi-container technologies.


\section*{Acknowledgements}
We would like to thank the staff of the Optimization Research Group of NICTA (National ICT of 
Australia) for allowing us to use Chuffed and G12/Gu\-ro\-bi solvers, as well as for granting us the 
computational resources needed for building and testing \solver{}. Thanks also to Michael Veksler 
for giving us an updated version of the HaifaCSP solver.

\bibliographystyle{named}
\bibliography{biblio}

\begin{thebibliography}{}

\bibitem[\protect\citeauthoryear{Amadini and Stuckey}{2014}]{paper_cp}
Roberto Amadini and Peter~J. Stuckey.
\newblock {Sequential Time Splitting and Bounds Communication for a Portfolio
  of Optimization Solvers}.
\newblock In {\em CP}, 2014.

\bibitem[\protect\citeauthoryear{Amadini \bgroup \em et al.\egroup
  }{2014a}]{mzn2feat}
Roberto Amadini, Maurizio Gabbrielli, and Jacopo Mauro.
\newblock {An Enhanced Features Extractor for a Portfolio of Constraint
  Solvers}.
\newblock In {\em SAC}, 2014.

\bibitem[\protect\citeauthoryear{Amadini \bgroup \em et al.\egroup
  }{2014b}]{paper_lion}
Roberto Amadini, Maurizio Gabbrielli, and Jacopo Mauro.
\newblock {Portfolio Approaches for Constraint Optimization Problems}.
\newblock In {\em LION}, 2014.

\bibitem[\protect\citeauthoryear{Amadini \bgroup \em et al.\egroup
  }{2014c}]{sunny}
Roberto Amadini, Maurizio Gabbrielli, and Jacopo Mauro.
\newblock {SUNNY: a Lazy Portfolio Approach for Constraint Solving}.
\newblock {\em {TPLP}}, 2014.

\bibitem[\protect\citeauthoryear{Amadini \bgroup \em et al.\egroup
  }{2015}]{sunnycp}
Roberto Amadini, Maurizio Gabbrielli, and Jacopo Mauro.
\newblock {SUNNY-CP: a Sequential CP Portfolio Solver}.
\newblock In {\em SAC}, 2015.

\bibitem[\protect\citeauthoryear{Brucker \bgroup \em et al.\egroup
  }{1999}]{brucker1999resource}
Peter Brucker, Andreas Drexl, Rolf M{\"o}hring, Klaus Neumann, and Erwin Pesch.
\newblock {Resource-constrained project scheduling: Notation, classification,
  models, and methods}.
\newblock {\em European journal of operational research}, 1999.

\bibitem[\protect\citeauthoryear{Cenamor \bgroup \em et al.\egroup
  }{2014}]{ibacop}
Isabel Cenamor, Tom{\'a}s de~la Rosa, and Fernando Fern{\'a}ndez.
\newblock {IBACOP and IBACOP2 Planner}, 2014.

\bibitem[\protect\citeauthoryear{Cire \bgroup \em et al.\egroup
  }{2014}]{cire2014parallel}
Andre Cire, Serdar Kadioglu, and Meinolf Sellmann.
\newblock Parallel restarted search.
\newblock In {\em AAAI}, 2014.

\bibitem[\protect\citeauthoryear{Duda \bgroup \em et al.\egroup }{2000}]{knn}
Richard~O. Duda, Peter~E. Hart, and David~G. Stork.
\newblock {\em Pattern Classification (2Nd Edition)}.
\newblock Wiley-Interscience, 2000.

\bibitem[\protect\citeauthoryear{Gomes and
  Selman}{2001}]{DBLP:journals/ai/GomesS01}
Carla~P. Gomes and Bart Selman.
\newblock Algorithm portfolios.
\newblock {\em Artif. Intell.}, 2001.

\bibitem[\protect\citeauthoryear{Hoos \bgroup \em et al.\egroup
  }{2014}]{claspfolio}
Holger Hoos, Marius~Thomas Lindauer, and Torsten Schaub.
\newblock {claspfolio 2: Advances in Algorithm Selection for Answer Set
  Programming}.
\newblock {\em {TPLP}}, 2014.

\bibitem[\protect\citeauthoryear{Hoos \bgroup \em et al.\egroup
  }{2015a}]{aspeed}
Holger~H. Hoos, Roland Kaminski, Marius~Thomas Lindauer, and Torsten Schaub.
\newblock {aspeed: Solver scheduling via answer set programming}.
\newblock {\em {TPLP}}, 2015.

\bibitem[\protect\citeauthoryear{Hoos \bgroup \em et al.\egroup
  }{2015b}]{lindauer-lion15a}
Holger~H. Hoos, Marius~Thomas Lindauer, and Frank Hutter.
\newblock {From Sequential Algorithm Selection to Parallel Portfolio
  Selection}.
\newblock In {\em LION}, 2015.

\bibitem[\protect\citeauthoryear{Hurley \bgroup \em et al.\egroup
  }{2014}]{proteus}
Barry Hurley, Lars Kotthoff, Yuri Malitsky, and Barry O'Sullivan.
\newblock {Proteus: A Hierarchical Portfolio of Solvers and Transformations}.
\newblock In {\em CPAIOR}, 2014.

\bibitem[\protect\citeauthoryear{Hutter \bgroup \em et al.\egroup
  }{2011}]{smac}
Frank Hutter, Holger~H. Hoos, and Kevin Leyton{-}Brown.
\newblock {Sequential Model-Based Optimization for General Algorithm
  Configuration}.
\newblock In {\em LION}, 2011.

\bibitem[\protect\citeauthoryear{Kadioglu \bgroup \em et al.\egroup
  }{2010}]{isac}
Serdar Kadioglu, Yuri Malitsky, Meinolf Sellmann, and Kevin Tierney.
\newblock {ISAC - Instance-Specific Algorithm Configuration}.
\newblock In {\em ECAI}, 2010.

\bibitem[\protect\citeauthoryear{Kadioglu \bgroup \em et al.\egroup
  }{2011}]{3s}
Serdar Kadioglu, Yuri Malitsky, Ashish Sabharwal, Horst Samulowitz, and Meinolf
  Sellmann.
\newblock {Algorithm Selection and Scheduling}.
\newblock In {\em CP}, 2011.

\bibitem[\protect\citeauthoryear{Malitsky \bgroup \em et al.\egroup
  }{2012}]{3s-par}
Yuri Malitsky, Ashish Sabharwal, Horst Samulowitz, and Meinolf Sellmann.
\newblock {Parallel {SAT} Solver Selection and Scheduling}.
\newblock In {\em CP}, 2012.

\bibitem[\protect\citeauthoryear{Malitsky \bgroup \em et al.\egroup
  }{2013}]{cshc}
Yuri Malitsky, Ashish Sabharwal, Horst Samulowitz, and Meinolf Sellmann.
\newblock {Algorithm Portfolios Based on Cost-Sensitive Hierarchical
  Clustering}.
\newblock In {\em IJCAI}, 2013.

\bibitem[\protect\citeauthoryear{Nethercote \bgroup \em et al.\egroup
  }{2007}]{Nethercote07minizinc:towards}
Nicholas Nethercote, Peter~J. Stuckey, Ralph Becket, Sebastian Brand,
  Gregory~J. Duck, and Guido Tack.
\newblock {MiniZinc: Towards a Standard CP Modelling Language}.
\newblock In {\em CP}, 2007.

\bibitem[\protect\citeauthoryear{O'Mahony \bgroup \em et al.\egroup
  }{2008}]{cphydra}
Eoin O'Mahony, Emmanuel Hebrard, Alan Holland, Conor Nugent, and Barry
  O'Sullivan.
\newblock Using case-based reasoning in an algorithm portfolio for constraint
  solving.
\newblock {\em AICS}, 2008.

\bibitem[\protect\citeauthoryear{Rice}{1976}]{rice_algorithm_selection}
John~R. Rice.
\newblock {The Algorithm Selection Problem}.
\newblock {\em Advances in Computers}, 1976.

\bibitem[\protect\citeauthoryear{Rossi \bgroup \em et al.\egroup
  }{2006}]{handbook}
F.~Rossi, P.~van Beek, and T.~Walsh, editors.
\newblock {\em Handbook of Constraint Programming}.
\newblock Elsevier, 2006.

\bibitem[\protect\citeauthoryear{Roussel}{2011}]{ppfolio}
Olivier Roussel.
\newblock {ppfolio portfolio solver}.
\newblock http://www.cril.univ-artois.fr/\~{}roussel/ppfolio/, 2011.

\bibitem[\protect\citeauthoryear{Stuckey \bgroup \em et al.\egroup
  }{2010}]{DBLP:journals/constraints/StuckeyBF10}
Peter~J. Stuckey, Ralph Becket, and Julien Fischer.
\newblock {Philosophy of the MiniZinc challenge}.
\newblock {\em Constraints}, 2010.

\bibitem[\protect\citeauthoryear{Valenzano \bgroup \em et al.\egroup
  }{2012}]{arvandherd}
Richard~Anthony Valenzano, Hootan Nakhost, Martin M{\"{u}}ller, Jonathan
  Schaeffer, and Nathan~R. Sturtevant.
\newblock {ArvandHerd: Parallel Planning with a Portfolio}.
\newblock In {\em ECAI}, 2012.

\bibitem[\protect\citeauthoryear{Xu \bgroup \em et al.\egroup
  }{2008}]{satzilla}
Lin Xu, Frank Hutter, Holger~H. Hoos, and Kevin Leyton{-}Brown.
\newblock {SATzilla: Portfolio-based Algorithm Selection for SAT}.
\newblock {\em JAIR}, 2008.

\bibitem[\protect\citeauthoryear{Yun and Epstein}{2012}]{parrallelCSPPortfolio}
Xi~Yun and Susan~L. Epstein.
\newblock {Learning Algorithm Portfolios for Parallel Execution}.
\newblock In {\em LION}, 2012.

\end{thebibliography}

\end{document}